\documentclass[11pt]{article}

\usepackage[final]{acl}

\usepackage{times}
\usepackage{latexsym}
\usepackage[T2A,T1]{fontenc}
\usepackage[utf8]{inputenc}
\newcommand{\cyrtext}[1]{%
  {\fontencoding{T2A}\fontfamily{cmr}\selectfont #1}%
}
\usepackage{microtype}
\usepackage{inconsolata}
\usepackage{graphicx}

\usepackage{amsmath}
\usepackage{amssymb}      
\usepackage{booktabs}     
\usepackage{multirow}     
\usepackage{pifont}       
\usepackage{xcolor}

\newcommand{\rev}[1]{#1}

\newcommand{\new}[1]{#1}

\title{
Lost in Speech: Trilingual Spoken Hallucination Detection Across Audio and Transcripts
}

\author{
\textbf{Meruyert Aristombayeva\textsuperscript{1,*}},
\textbf{Jason S. Lucas\textsuperscript{2}},
\textbf{Chaewan Chun\textsuperscript{3}},
\textbf{Dongwon Lee\textsuperscript{3}}
\\
\textsuperscript{1}Satbayev University, Almaty, Kazakhstan \\
\textsuperscript{2}University of Colorado Boulder, Boulder, CO, USA \\
\textsuperscript{3}The Pennsylvania State University, University Park, PA, USA
\\
\small{
\textsuperscript{*}\textbf{Corresponding author:}
\href{mailto:m.aristombayeva@satbayev.university}
{m.aristombayeva@satbayev.university}
}
}

\begin{document}
\maketitle

\begin{abstract}
While text-based hallucination detection is well studied, reference-free detection of factual alterations in speech remains underexplored, especially for low-resource languages. Our spoken benchmark comprises 12,013 English, Russian, and Kazakh news samples with three synthetic alteration types and three severity levels, pairing source articles with rewrites as text, synthesized audio, and ASR transcripts. We add 290 fact-checked misinformation items collected in Russian (225) and Kazakh (65), translated into the other language and rendered through the same TTS–ASR pipeline. We evaluate fine-tuned multilingual encoders and zero-shot multimodal decoders on text, transcripts, and audio. Detectors receive only target inputs without source articles or external evidence; the task evaluates reference-free classification rather than evidence-grounded verification. Encoder degradation from source text to transcripts generally tracks per-language ASR error on the binary task. Among decoders, only Gemma-3n exceeds the binary majority-class baseline in macro-F1, on transcripts only; the other four do not exceed their respective baselines. Comparisons between audio and transcripts are
confounded by differences in evaluation coverage
and class balance. Synthetic-trained detectors achieve 0.82–0.88 macro-F1 on real-world misinformation source text; Russian provenance analysis reveals veracity-related and model-dependent machine-style signals, a key confound in synthetic hallucination benchmarks.
\end{abstract}

\section{Introduction}
In recent years, large neural models for language and speech have
demonstrated impressive generative abilities, yet they exhibit a troubling
propensity to hallucinate, producing information that is not grounded in
factual sources \citep{Frieske2024}. In the spoken setting the stakes are
higher still: transcription systems can emit entirely fabricated phrases, a
non-trivial fraction of which carry explicit real-world harms
\citep{koenecke2024careless}. Despite this, hallucination detection has been
studied almost exclusively in text \citep{ji2023survey}, while work in the
spoken domain remains scarce---and scarcer still for low-resource
languages.

Unlike textual inputs, spoken content is processed through multi-stage
pipelines that couple text-to-speech (TTS) synthesis and automatic speech
recognition (ASR), introducing noise and cascading error propagation. The
robustness of hallucination detection under realistic speech-pipeline
conditions is therefore still insufficiently understood
\citep{Atwany2025}. As large language and audio-language models are
deployed ever more widely, evaluating hallucination detection under
multilingual speech constraints has become increasingly critical
\citep{Frieske2024,Latif2023}.

\rev{Three} gaps remain open. First, dedicated benchmarks for hallucination in
audio and audio-language models have only begun to emerge, and they are
overwhelmingly English and question-answering oriented
\citep{kuan2024object,cheng2025ahabench}. Second, the few multilingual
hallucination benchmarks are text-only \citep{halluverse25,mushroom2025},
and recent evidence indicates that hallucination behavior does not track a
language's digital footprint in any simple way
\citep{obaidulislam2025multilingual}---motivating the study of a genuinely
low-resource language such as Kazakh alongside higher-resource English and
Russian. \rev{Third, existing hallucination benchmarks---synthetic by
construction---have not been validated against real-world misinformation:
whether detectors trained on LLM-generated hallucinations transfer to
human-written false content remains untested, a question sharpened by the
fact that in synthetic benchmarks text provenance (human vs.\ LLM) is
perfectly correlated with the label.} A further bottleneck is the lack of
large, balanced, and annotated
audio datasets, which is acute precisely where both TTS and ASR are least
mature \citep{Chun2025}. To our knowledge, no prior resource combines spoken
input, a \emph{detection} task, three languages including Kazakh, and
\emph{controlled} hallucination types at graded severity; existing work
covers only subsets of these axes.

\begin{figure*}[t]
  \centering
  \includegraphics[
    width=0.90\textwidth,
    trim={0cm 3cm 0cm 4cm},
    clip
  ]{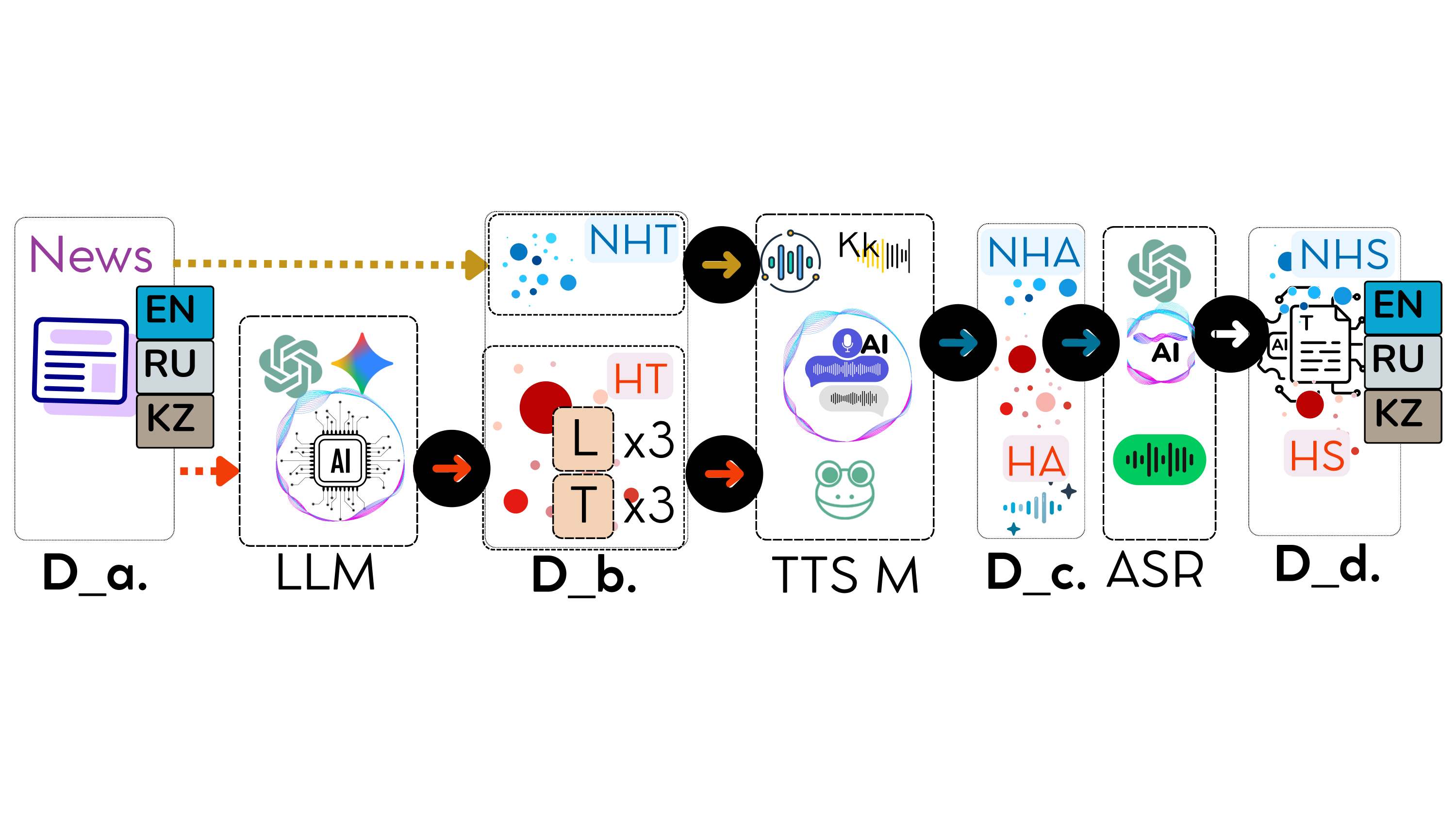}
  \caption{Data generation pipeline framework.
    \textbf{Legend:}
    $D_a$ = Original Data, $D_b$ = Generated Hallucinated Text,
    $D_c$ = Audio Data, $D_d$ = Transcribed Data;
    T = Type, L = Severity Level;
    NHT = Non-Hallucinated Text, HT = Hallucinated Text,
    NHA = Non-Hallucinated Audio, HA = Hallucinated Audio,
    NHS = Non-Hallucinated Speech, HS = Hallucinated Speech;
    LLM = Large Language Model, TTS\,M = Text-to-Speech Model,
    ASR = Automatic Speech Recognition\rev{; MT = Machine Translation}.}
  \label{fig:pipeline}
\end{figure*}

\rev{For benchmark construction, we operationally define a \emph{spoken
hallucination} as a synthetically introduced factual or contextual
alteration in a news article that is unsupported by, or contradictory to,
its source article. The source article is used to generate and assign the
benchmark labels but is not provided to the evaluated detectors. Thus, our experiments concern reference-free classification of
benchmark-defined, source-relative alterations rather than
evidence-grounded verification of factual truth.} To study this problem under
controlled yet realistic conditions, we construct a multilingual benchmark\footnote{%
\url{https://huggingface.co/datasets/maristombayeva/lost-in-speech}}
spanning English (high-resource), Russian (medium-resource), and Kazakh
(low-resource). Each sample pairs a reference article with a systematically
generated hallucinated counterpart of controlled hallucination type
(fabrication, contradiction, context inconsistency) and severity level
(mild, moderate, severe), while preserving topic and style. Our controlled
generation and LLM-as-a-judge validation descend from synthetic-hallucination
pipelines developed for text \citep{mishra2024fava,xie2024controlled}. To
simulate deployment, every version is provided in both text and speech form:
hallucinated texts are synthesized with language-specific TTS systems and
transcribed back with ASR (Fig.~\ref{fig:pipeline}). \rev{This design enables evaluation on clean text, ASR-transcribed
text, and raw audio, and supports analysis of how cascading
speech-pipeline noise---a documented risk in spoken language understanding
\citep{avila2023multimodal}---affects reference-free classification of
the benchmark labels.} We evaluate two model families on the resulting data: fine-tuned
multilingual encoders and zero-shot multimodal (audio-language) decoders.
Our experiments show TTS--ASR pipeline degradation
most pronounced for Kazakh.
\paragraph{Contributions.}
\begin{enumerate}
  \setlength{\itemsep}{1pt}
 \item \rev{A multilingual spoken benchmark for reference-free
classification of synthetically introduced factual and contextual
alterations, with controlled alteration types and graded severity:
12{,}013 news samples comprising source articles and aligned rewritten
counterparts in English, Russian, and Kazakh, represented as text,
synthesized speech, and ASR transcripts.}
  \item A structured LLM-based hallucination generation framework with
generator self-assessment and external validation by two independent cross-family LLM judges, supplemented by targeted human inspection of synthesized speech.
  \item \rev{The first real-world spoken evaluation split for this task:
  290 fact-checked fake news stories from factcheck.kz, collected
  natively in Russian (225) and Kazakh (65), each paired with a
  translation into the other language and with matched truthful
  negatives, rendered through the same TTS$\rightarrow$ASR pipeline
  (Section~\ref{ssec:realworld}).}
  \item A \rev{Russian provenance analysis} that probes factuality and
human-vs-machine text signals---a confound inherent to synthetic
benchmarks---by evaluating detectors on human-written false content
  (Section~\ref{ssec:rq3}).
  \item A language-specific TTS$\rightarrow$ASR pipeline with
  best-of-breed per-language ASR, and a comprehensive empirical study of
  fine-tuned encoders and zero-shot audio-language decoders across text
  and audio settings, with ASR error analysis \new{and an analysis of
decoder failure modes, output parseability, and the input-length limits
of audio encoders}, addressing RQ1--RQ3
  (Section~\ref{sec:eval}).
\end{enumerate}

\section{Related Work}

\noindent\textbf{Hallucination detection in text and speech.}
Text-based hallucination research spans error taxonomies
\citep{ji2023survey,Huang2023}, synthetic supervision
\citep{mishra2024fava,xie2024controlled}, and reference-free detection
through sampling consistency \citep{manakul2023selfcheckgpt}.
Speech studies examine model-generated errors
\citep{Frieske2024,Atwany2025} and associated harms
\citep{koenecke2024careless}, while audio benchmarks assess object
hallucination and audio question answering
\citep{kuan2024object,cheng2025ahabench}. Multilingual hallucination
benchmarks \citep{halluverse25,mushroom2025,obaidulislam2025multilingual}
focus on text.

\noindent\textbf{Fact-checking and spoof detection.}
FEVER \citep{thorne2018fever}, AVeriTeC
\citep{schlichtkrull2023averitec}, and X-Fact \citep{gupta2021xfact}
support evidence-grounded claim verification. Our detectors predict
source-relative alteration labels without access to source articles
or external evidence. FakeSV \citep{qi2023fakesv} and MAD
\citep{Chun2025} address video misinformation and spoken dialogue,
respectively. Spoof detection
\citep{wang2024asvspoof5,yi2023audiodeepfake,zhang2023partialspoof}
targets speech authenticity rather than factual or contextual alterations.

\noindent\textbf{Positioning.}
Audio-language models \citep{chu2024qwen2audio} and ASR systems
\citep{radford2023whisper} enable direct audio and transcript-based
processing; AIR-Bench \citep{yang2024airbench} evaluates audio
comprehension and instruction-following. Kazakh resources support
recognition and synthesis
\citep{khassanov2021ksc,kazakhtts,mussakhojayeva2022kazakhtts2}
and knowledge evaluation \citep{togmanov2025kazmmlu}.
Our benchmark combines English, Russian, and Kazakh text, synthesized
speech, and ASR transcripts for reference-free classification of
controlled factual and contextual alterations by type and severity.
We also test real-world transfer and examine text provenance---the target
of machine-generated text detection \citep{wang2024m4gt}---as a potential
confound.

\section{\rev{Benchmark Construction}}
\label{sec:generation}

\subsection{\rev{Synthetic Subset}}
\label{ssec:data}

\rev{We constructed a corpus of 12,013 samples based on news articles
gathered from major Kazakhstani news platforms (e.g., nur.kz, forbes.kz,
tengrinews.kz, kapital.kz), covering diverse topics (politics, economy,
technology); each sample is either an original article or one of its
generated hallucinated versions (Table~\ref{tab:dataset}).} The articles span three languages---Kazakh, Russian, and English---and are treated as factual references since they originate from established outlets. \rev{We generated hallucinated versions with three hallucination types (fabrication, contradiction, and context inconsistency) and three severity levels (mild, moderate, severe), following \citet{Huang2023}: the English batches were produced with \texttt{gpt-3.5-turbo} and \texttt{gpt-4}, and the Russian and Kazakh batches with \texttt{gemini-2.0-flash-lite}; the prompts of the final pipeline, including all language variants, are documented in Appendix~\ref{app:prompts}.} Table~\ref{tab:dataset} summarizes the dataset distribution. Each sample is available as text, synthesized audio, and ASR transcript (Fig.~\ref{fig:pipeline}). As speech is produced via a TTS$\rightarrow$ASR pipeline, transcription errors reflect cascading noise from both synthesis and recognition components, particularly in low-resource Kazakh.

\begin{table}[t]
\caption{Dataset summary across languages.
Res = resource
 level (H = high, M = medium, L = low).
Type: C = contradiction, F = fabrication, I = inconsistency.
Severity: Mild, Moderate, Severe.
\rev{Rows marked ``real'' denote the real-world evaluation split from
factcheck.kz (Section~\ref{ssec:realworld});
$^{\dagger}$each language combines natively collected fakes (225 ru, 65
kz) with parallel translations of the other language's items. Real items
carry binary labels only and are used exclusively for evaluation. The
290 truthful items of each real row are original articles of the
synthetic subset above (not additional data); only the 290 fakes per
language are new items.}}
\label{tab:dataset}
\centering
\footnotesize
\renewcommand{\arraystretch}{1.02}
\setlength{\tabcolsep}{3pt}
\resizebox{\columnwidth}{!}{%
\begin{tabular}{l l r rr rrr rrr}
\toprule
\textbf{Lang} & \textbf{Res} & \textbf{Total} &
\multicolumn{2}{c}{\textbf{Binary}} &
\multicolumn{3}{c}{\textbf{Type}} &
\multicolumn{3}{c}{\textbf{Severity}} \\
\cmidrule(lr){4-5}
\cmidrule(lr){6-8}
\cmidrule(lr){9-11}
 &  &  &
\textbf{No} & \textbf{Yes} &
\textbf{C} & \textbf{F} & \textbf{I} &
\textbf{Mild} & \textbf{Mod} & \textbf{Sev} \\
\midrule
en & H & 3967 & 1340 & 2627 & 903 & 938 & 786 & 875 & 877 & 875 \\
kz & L & 3978 & 1113 & 2865 & 947 & 956 & 962 & 957 & 953 & 955 \\
ru & M & 4068 & 1259 & 2809 & 962 & 960 & 887 & 934 & 937 & 938 \\
\midrule
\rev{ru (real)$^{\dagger}$} & \rev{M} & \rev{580} & \rev{290} & \rev{290} & \rev{--} & \rev{--} & \rev{--} & \rev{--} & \rev{--} & \rev{--} \\
\rev{kz (real)$^{\dagger}$} & \rev{L} & \rev{580} & \rev{290} & \rev{290} & \rev{--} & \rev{--} & \rev{--} & \rev{--} & \rev{--} & \rev{--} \\
\bottomrule
\end{tabular}
}
\end{table}

To create diverse and controlled hallucinations, we implemented a category-based prompting framework inspired by factuality taxonomies in prior work \citep{Huang2023, lucas2023, nahar2024}. Each article was rewritten according to a predefined hallucination type and severity level.

\rev{We distinguish \textit{Factual Contradiction} (statements directly
conflicting with facts or information from the original article),
\textit{Factual Fabrication} (insertion of fabricated yet
plausible-sounding details not grounded in the source), and
\textit{Context Inconsistency} (subtle alterations that distort the
meaning, emphasis, or context of the original article without explicit
factual errors).}

Each generation was guided by a structured prompt incorporating the
original article, the target hallucination type, and severity level.
The generation prompt also required the generating model to provide a
structured self-assessment of its output along predefined quality
dimensions. Because such in-prompt self-assessment is not independent
of generation, we additionally conducted a separate external validation
stage using two judge models from families disjoint from the generators
(Section~\ref{ssec:quality}). All generation and evaluation prompts are
provided in Appendix~\ref{app:prompts}.

\subsection{\rev{Real-World Subset}}
\label{ssec:realworld}

\rev{Synthetic benchmarks carry an inherent confound: every hallucinated
sample is LLM-generated while every faithful sample is human-written, so
provenance is perfectly correlated with the label. To reduce this direct
human-versus-LLM label correlation and to test performance on naturally
occurring misinformation, we construct a real-world evaluation split from
factcheck.kz, a professional Kazakhstani fact-checking organization.}

\rev{We collected 290 news items verified as false by factcheck.kz,
downloaded from its public archive in October 2024: 225 published
natively in Russian and 65 natively in Kazakh (after removing 15 exact duplicates from
the initial Russian collection), spanning topics comparable to the
synthetic corpus (politics, society, health, technology). Each item
consists of the fake claim or article as it circulated, with the
fact-checking verdict as ground truth. To enable controlled cross-lingual
comparison on identical content, every item is additionally translated
into the other language using Google Translate, yielding a parallel
Russian--Kazakh fake-news corpus of 290 stories per language, of which 65
Kazakh and 225 Russian items are natively circulating misinformation. Six
claims were independently fact-checked in both languages; these
near-parallel native pairs are flagged in the release. Translation quality was manually
validated by a fluent bilingual author on all 290 translations, all of
which were judged adequate (Appendix~\ref{app:realworld}).}

\rev{As truthful negatives, we select 290 human-written original
articles per language from the synthetic subset, matched to the fakes
via greedy TF-IDF cosine similarity over topic and time period. These
articles are excluded from the training and development splits of every
model evaluated on the real-world data
(Appendix~\ref{app:training}), so no model evaluated on this split has
seen any of its items during training. All real-world items---fake and truthful---are rendered through the
identical TTS$\rightarrow$ASR pipeline described in
Section~\ref{ssec:speech}, so the real-world split was evaluated in the same
three modalities as the synthetic corpus (only URLs, labels, and metadata
are released; see Ethical Considerations). This split is used exclusively for evaluation; no model is trained
on it. Unlike the synthetic subset, real-world items carry only binary
labels, since fact-checkers do not annotate our type/severity taxonomy;
manual taxonomy alignment is left to future work.}

\subsection{Speech Synthesis and Transcription}
\label{ssec:speech}

We synthesized speech for both original and hallucinated texts using language-specific TTS systems: Coqui XTTS-v2 \citep{xttsv2} for English, Silero TTS \citep{SileroModels} for Russian, and an open-source Kazakh TTS model \citep{kazakhtts}. Multiple voices were employed to increase acoustic diversity. Due to limited availability of high-quality Kazakh TTS systems, synthesized Kazakh speech exhibited reduced naturalness and occasional phonetic instability. \rev{Hallucination labels are assigned at the text-generation stage
relative to the source article; subsequent TTS--ASR discrepancies are
treated as pipeline degradation and do not alter these labels.}

All audio samples were transcribed using ASR. English and Russian were processed using Whisper-large-v3, while Kazakh was transcribed using a fine-tuned wav2vec2-large model with post-processing for punctuation. We adopt the strongest available ASR per language: as shown in Table~\ref{tab:asr_choice} (Appendix~\ref{app:models}), Whisper-large-v3 is highly accurate on English and Russian but fails on low-resource Kazakh, whereas the fine-tuned wav2vec2 model nearly halves the Kazakh error rate on identical audio. All WER/CER values are computed at the corpus level after lowercasing, punctuation removal, and digit removal, applied identically across languages.

We quantify transcription distortion using WER/CER between source texts and ASR outputs: EN 7.30\%/2.74\%, RU 9.08\%/4.74\%, and KZ 34.04\%/16.90\% (overall WER 13.02\%, CER 6.46\%). The substantially higher Kazakh error rates reflect compounded degradation across the TTS--ASR cascade. In particular, phonetic distortions introduced by TTS in named entities and region-specific terms (e.g., Oskemen realized as Askemin) are faithfully transcribed by ASR but counted as full word errors against the original text. The news domain further amplifies this effect due to a high density of proper names and Kazakhstan-specific terminology (e.g., akimat), which are underrepresented in multilingual training corpora. Additionally, Kazakh's agglutinative morphology increases lexical variability, so even minor deviations often result in complete word-level mismatches, inflating WER relative to CER.

\rev{This best-of-breed choice minimizes, but does not eliminate, the
cross-language tooling gap; the residual gap (Kazakh WER remains roughly
four times higher than English and Russian) is examined by relating per-language ASR error
to downstream detection degradation (Sections~\ref{ssec:orig-vs-trans}~and~\ref{ssec:audio-vs-trans}).}

\subsection{Quality Evaluation: LLM vs. Human}
\label{ssec:quality}

To validate the quality of both raw hallucinated texts and synthesized speech, we perform two tasks.

\paragraph{Evaluating Hallucinated Texts.}
We adopted an LLM-as-a-Judge framework \citep{Zheng2023, Li2024, Gu2025}.
Generation prompts included an in-prompt self-assessment; independent
validation used a stratified sample of 216 articles ($n=8$ per
type$\times$severity$\times$language cell) and two external judges,
Claude Haiku 4.5 and DeepSeek Chat, under the same protocol.

\rev{The evaluation considered nine dimensions, applied type-dependently
(seven per type): Adequacy of Type (AT), Fluency (Fl), Coherence (Co),
Logical Coherence (LC), Internal Consistency (IC), Credibility (Cr),
Factual Accuracy (FA), Relevance (Re), and Satisfaction (Sa); definitions
are in Appendix~\ref{app:prompts}.}

Each criterion was rated on a five-point ordinal scale from \textit{Very Low (1)} to \textit{Very High (5)}. Table~\ref{tab:llm_eval}
\rev{(Appendix~\ref{app:quality})} reports mean Claude scores on this
sample by type, severity, and language. As the scale is ordinal, means
serve only as descriptive indicators.

\rev{For both factuality-bearing
types, the judge assigns low credibility and factual accuracy already at
mild severity ($\sim$1.2--2.0) and floor scores ($\approx$1.0) at severe
levels, confirming that the intended factual degradation is realized. Adequacy of Type and Satisfaction increase with
severity for all three types, reaching 4.9--5.0 at the severe level, so
generated articles match their target category and level most clearly
where distortion is strongest. Fluency declines only
moderately with severity and does not collapse for any type (severe-level
means of 2.6--4.0 for contradictions; Table~\ref{tab:llm_eval}),
indicating that the framework manipulates semantic grounding rather than
linguistic form.}

\paragraph{Cross-Judge Validation.}
A natural concern is that the quality scores in Table~\ref{tab:llm_eval} could reflect
judge-specific idiosyncrasies. To test this, we measured agreement between the two
independent external judges, Claude Haiku 4.5 and DeepSeek Chat, on the same stratified
sample and per quality dimension (Table~\ref{tab:judge_agreement},
Appendix~\ref{app:quality}).
Agreement is high on the two factuality-oriented dimensions that underlie the
hallucination signal---Factual Accuracy and Credibility (quadratic-weighted
$\kappa = 0.875$ and $0.897$; Spearman $\rho = 0.91$ and $0.92$)---with the two judges
agreeing within one point on $100\%$ of these items. For the remaining, more subjective
dimensions, raw agreement remains high (within one point on $71$--$97\%$ of items),
although weighted $\kappa$ is lower for the most concentrated scales (e.g., Internal
Consistency, Logical Coherence), where limited score variance deflates chance-corrected
agreement. Overall, the cross-judge results corroborate the trends in
Table~\ref{tab:llm_eval}, indicating that the reported quality patterns are robust
across the two external evaluators.

\paragraph{Evaluating Synthesized Speech.}
Given Kazakh's high WER/CER, a fluent native Kazakh-speaking author
randomly selected and evaluated 120 audio samples using predefined
criteria for intelligibility, faithfulness to the source text, and
exclusion due to audio quality.

The results indicate that 94.2\% of the samples were rated as mostly intelligible and 5.8\% as poorly intelligible, with no samples classified as fully unintelligible. Regarding semantic fidelity, 61.7\% exhibited minor discrepancies and 38.3\% major discrepancies relative to the source text. No samples were excluded due to critical audio quality issues. These findings confirm that the Kazakh pipeline introduces noticeable semantic distortion while maintaining general perceptual comprehensibility, thereby motivating focused robustness analysis in the low-resource setting.

\section{Empirical Evaluation}
\label{sec:eval}

We aim to answer the following research questions:
\textit{RQ1: How robust are state-of-the-art multilingual encoder models
for reference-free classification of synthetically introduced factual
alterations when evaluated on original text versus ASR-transcribed text?}
\textit{RQ2: Under identical zero-shot prompting conditions, can multimodal
decoder models perform reference-free classification from ASR transcripts and
from direct audio, and are the two inputs comparable on this benchmark?}
\rev{\textit{RQ3: Do classifiers trained on synthetically altered texts
transfer to real-world, human-written misinformation, and to what extent
is their binary signal attributable to human-vs-machine text
discrimination rather than veracity-related content?}}

\rev{To address RQ1, a model receives only the target news item,
represented either as text ($D_a$) or as an ASR transcript ($D_d$), and
predicts labels assigned during controlled dataset construction. The
corresponding source article and external evidence are not provided at
inference time. We evaluate binary classification (unaltered vs.\
synthetically altered) and multiclass classification of the assigned
alteration type and severity. Consequently, these experiments measure
reference-free recognition of benchmark-defined alterations and should
not be interpreted as evidence-grounded factual verification.} We compare (i) fine-tuned multilingual encoder-based transformers and (ii) zero-shot in-context learning with decoder-based architectures. The models and modality support are summarized in Table~\ref{tab:model_support} \rev{(Appendix~\ref{app:models})}.

\rev{Throughout, we report accuracy and macro-F1; training details are
given in Appendix~\ref{app:training}.}

\subsection{RQ1 $\rightarrow$ Text-based: Original Text vs. Transcripts}
\label{ssec:orig-vs-trans}

We assess multilingual encoder robustness in text-only deployment scenarios. Specifically, we compare performance on original articles ($D_a$) and their ASR-transcribed counterparts ($D_d$).

Table~\ref{tab:encoder_performance} shows that fine-tuned encoders achieve
strong binary hallucination detection performance (F1: 0.52--0.89), while
fine-grained type (F1: 0.14--0.68) and severity classification
(F1: 0.15--0.66) remain more challenging. Performance is predominantly
higher on original text than on ASR-transcribed content, though not
uniformly so: in a few language--task cells transcripts match or exceed
original text (e.g., XLM-R (base) on Russian). mDeBERTa exhibits the most
robust cross-lingual performance; ReMBERT is competitive on binary
detection but performs poorly on type and severity tasks (F1 .153--.188).
For mDeBERTa and ReMBERT, binary-task degradation from $D_a$ to $D_d$
increases with per-language ASR error, whereas this trend is inconsistent
for type and severity tasks and XLM-R models.

\begin{table}[t]
\caption{Finetuned encoder performance (Acc/F1).
Task: B = Binary, T = Type, L = Level;
Split: O = original text, T = ASR-transcribed text.
Bold indicates the best F1 within each row (among the four encoders).}
\label{tab:encoder_performance}
\centering
\footnotesize
\renewcommand{\arraystretch}{1.05}
\setlength{\tabcolsep}{3pt}
\resizebox{\columnwidth}{!}{
\begin{tabular}{@{}l l c *{4}{c}@{}}
\toprule
\textbf{Task} & \textbf{Lan} & \textbf{Split} &
\textbf{XLM-R (b)} & \textbf{XLM-R (l)} & \textbf{mDeBERTa} & \textbf{ReMBERT} \\
\cmidrule(lr){4-7}
& & & \multicolumn{4}{c}{\textit{Acc/F1}} \\
\midrule
 & en & O & .826/.828 & \textbf{.865/.863} & .843/.843 & .879/.860 \\
                &    & T & .838/.832 & .658/.522 & \textbf{.840/.841} & .834/.815 \\
\cmidrule(lr){2-7}
\textbf{B}      & kz & O & .870/.873 & .841/.846 & \textbf{.893/.893} & .910/.876 \\
                &    & T & .826/.826 & .711/.591 & \textbf{.875/.874} & .802/.782 \\
\cmidrule(lr){2-7}
                & ru & O & .809/.814 & .738/.747 & .838/.841 & \textbf{.862/.845} \\
                &    & T & \textbf{.832/.835} & .703/.580 & .824/.827 & .788/.767 \\
\midrule
  & en & O & .612/.599 & \textbf{.680/.679} & .600/.596 & .388/.187 \\
                &    & T & .571/.564 & \textbf{.673/.666} & .627/.623 & .331/.166 \\
\cmidrule(lr){2-7}
\textbf{T}      & kz & O & .612/.606 & .675/.674 & \textbf{.681/.676} & .306/.156 \\
                &    & T & .260/.138 & \textbf{.635/.636} & .559/.560 & .347/.172 \\
\cmidrule(lr){2-7}
                & ru & O & .669/.663 & \textbf{.665/.665} & .616/.616 & .345/.171 \\
                &    & T & \textbf{.676/.677} & .665/.660 & .645/.644 & .353/.174 \\
\midrule
 & en & O & .596/.590 & \textbf{.616/.611} & .604/.596 & .344/.171 \\
                &    & T & \textbf{.633/.616} & .290/.150 & .568/.570 & .394/.188 \\
\cmidrule(lr){2-7}
\textbf{L}      & kz & O & .592/.592 & \textbf{.660/.660} & .664/.656 & .326/.163 \\
                &    & T & .559/.555 & .374/.182 & \textbf{.601/.586} & .297/.153 \\
\cmidrule(lr){2-7}
                & ru & O & .624/.624 & \textbf{.668/.655} & .598/.595 & .336/.168 \\
                &    & T & .618/.611 & .333/.167 & \textbf{.633/.634} & .316/.160 \\
\bottomrule
\end{tabular}
}
\end{table}

\subsection{RQ2 $\rightarrow$ Audio-Based: Decoders on Audio and Transcripts}
\label{ssec:audio-vs-trans}

\new{We evaluate five decoder models (1.5B--8B) in a zero-shot setting,
reporting ASR-transcript ($D_d$) results for all five and direct-audio
($D_c$) results for three (Table~\ref{tab:decoder_classification}).
We report direct prompting only. Under chain-of-thought (CoT) prompting,
Qwen2-Audio requested audio in every fine-grained audio call,
Qwen2.5-Omni refused 33\% of Kazakh binary audio calls
(versus 2\% under direct prompting), and up to 78\% of Gemma-3n's
responses were unparsable. Coverage differs across cells (see caption),
limiting direct comparisons.}

\new{Near-chance scores reflect degenerate output rather than partial
competence, and the mechanism differs by model. Qwen2.5-Omni parses over 96\%
of transcript calls but rarely asserts an alteration: on Kazakh its binary
predictions are constant negative---accuracy .279 on the full corpus, matching
the unaltered share up to unparsed items, and .834 on the 344-item audio subset,
where the unaltered share is .854 and the remaining 2\% are refusals counted
as errors. Qwen2-Audio and Step-Audio-2-mini are also constant on the balanced
binary subsets (.500/.333 in all languages), though Qwen2-Audio exceeds the
fine-grained macro-F1 baseline on transcripts (.198--.357) and Step-Audio-2-mini on
English type (.467/.401). LFM2-Audio, documented for English only, fails at the
output level. Gemma-3n is the only decoder above the binary baseline on
transcripts in all three languages, with macro-F1 .613/.581/.630 against
.40--.42, and it also exceeds the severity baseline (.399/.328/.370 against
.167); its type predictions stay near baseline.}

\new{We do not compare modalities. Audio items have a median duration of
96\,s (English: 212\,s; 98\% exceed 30\,s). Qwen2-Audio's Whisper-large-v3
front end is documented to truncate beyond 30\,s; Qwen2.5-Omni uses a
block-wise streaming encoder and LFM2-Audio a FastConformer encoder, neither
documented with that limit. We did not verify per-model truncation in our
inference code, so under direct audio a model may receive only part of an
article labelled over its full text. The audio subsets are also
skewed---the Kazakh one is 85\% unaltered against 28\% in the corpus---so
their accuracies are not comparable to corpus baselines. For Kazakh, 38\% of
inspected audio also shows major semantic discrepancies from the source text
(Section~\ref{ssec:quality}), so the text-level label may not be recoverable
from audio at all. The near-constant
audio predictions are consistent with these mismatches and say nothing about
audio reasoning.}

\new{LFM2-Audio and Step-Audio-2-mini are evaluated outside their
documented language coverage on Russian and Kazakh by design, to show
where open audio-language models stop working.}

\begin{table*}[t]
\caption{Decoder performance on direct audio (A) and ASR transcripts (T).
Direct prompting only; chain-of-thought output was unusable
(Section~\ref{ssec:audio-vs-trans}). Acc and macro-F1 use the same items;
unparsable and \emph{none} responses count as errors. Cells with different $n$
are not comparable. Transcript $n$: full corpus for $\bullet$,
$\blacktriangle$, $\blacksquare$ (3{,}965/3{,}977/4{,}067 binary;
2{,}627/2{,}865/2{,}809 fine-grained, en/kz/ru); 100/150 for $\blacklozenge$;
60/90 for $\star$. Audio $n$: full corpus for $\bullet$ on en, 344/150 kz,
100/150 ru; 100/150 for $\blacktriangle$; 60/90 for $\blacklozenge$; $\star$
transcripts only, $\blacksquare$ audio excluded (raw responses not retained)
(--). Majority baselines (Acc/macro-F1): .663/.399, .720/.419, .691/.409
(full binary corpus, en/kz/ru); .500/.333 balanced binary subsets;
full-corpus type .357/.176, .336/.168, .343/.170 and severity
.334/.167 (en/kz/ru); .333/.167 on the balanced fine-grained subsets. \textbf{Bold}: best among the full-corpus models ($\bullet$,
$\blacktriangle$, $\blacksquare$) on transcripts, the only comparable cells.
$^{\dag}$audio not delivered in 7--13\% of calls; not interpretable.
M: $\blacklozenge$ = LFM2-Audio-1.5B, $\blacktriangle$ =
Qwen2-Audio-7B-Instruct, $\bullet$ = Qwen2.5-Omni-3B, $\star$ =
Step-Audio-2-mini, $\blacksquare$ = Gemma-3n-E4B-it.}
\label{tab:decoder_classification}
\centering
\footnotesize
\renewcommand{\arraystretch}{1.05}
\setlength{\tabcolsep}{6pt}
\begin{tabular}{cl cc cc cc}
\toprule
& &
\multicolumn{2}{c}{\textbf{Binary}} &
\multicolumn{2}{c}{\textbf{Type}} &
\multicolumn{2}{c}{\textbf{Level}} \\
\cmidrule(lr){3-4} \cmidrule(lr){5-6} \cmidrule(lr){7-8}
\textbf{M} & \textbf{Lang} &
\textbf{A} & \textbf{T} &
\textbf{A} & \textbf{T} &
\textbf{A} & \textbf{T} \\
\cmidrule(lr){3-3} \cmidrule(lr){4-4} \cmidrule(lr){5-5} \cmidrule(lr){6-6} \cmidrule(lr){7-7} \cmidrule(lr){8-8}
& & \multicolumn{6}{c}{\scriptsize(Acc / F1)} \\
\midrule
\multirow{3}{*}{$\blacklozenge$}
 & en & .250/.336 & .140/.186 & .000/.000 & .013/.024 & .000/.000 & .080/.114 \\
 & kz & .183/.254$^{\dag}$ & .190/.254 & .000/.000 & .013/.024 & .000/.000 & .000/.000 \\
 & ru & .067/.094$^{\dag}$ & .170/.207 & .011/.020 & .013/.025 & .000/.000 & .007/.013 \\
\midrule
\multirow{3}{*}{$\blacktriangle$}
 & en & .500/.333 & .391/.339 & .333/.167 & \textbf{.313/.207} & .000/.000 & .423/.357 \\
 & kz & .500/.333 & .294/.250 & .333/.167 & \textbf{.328/.210} & .000/.000 & .327/.280 \\
 & ru & .500/.333 & .346/.296 & .333/.167 & \textbf{.286/.198} & .000/.000 & .280/.289 \\
\midrule
\multirow{3}{*}{$\bullet$}
 & en & .364/.295 & .359/.288 & .032/.058 & .024/.047 & .106/.148 & .146/.182 \\
 & kz & .834/.455 & .279/.219 & .000/.000 & .030/.050 & .013/.025 & .021/.036 \\
 & ru & .510/.355 & .339/.285 & .013/.026 & .038/.070 & .053/.084 & .088/.122 \\
\midrule
\multirow{3}{*}{$\star$}
 & en & -- & .500/.333 & -- & .467/.401 & -- & .400/.304 \\
 & kz & -- & .500/.333 & -- & .333/.168 & -- & .333/.167 \\
 & ru & -- & .500/.333 & -- & .333/.277 & -- & .278/.211 \\
\midrule
\multirow{3}{*}{$\blacksquare$}
 & en & -- & \textbf{.683/.613} & -- & .311/.165 & -- & \textbf{.468/.399} \\
 & kz & -- & \textbf{.607/.581} & -- & .262/.154 & -- & \textbf{.340/.328} \\
 & ru & -- & \textbf{.656/.630} & -- & .256/.173 & -- & \textbf{.403/.370} \\
\bottomrule
\end{tabular}
\end{table*}

\subsection{\rev{RQ3 $\rightarrow$ Real-World Transfer and the Provenance Confound}}
\label{ssec:rq3}

\rev{\paragraph{Real-world transfer.} We take the strongest fine-tuned
encoders from Section~\ref{ssec:orig-vs-trans} (ReMBERT and mDeBERTa),
trained exclusively on the synthetic subset, and evaluate them on the
real-world split (Table~\ref{tab:realworld}). On original text, binary
macro-F1 on real-world data matches or exceeds synthetic test
performance for both encoders and both languages (e.g., ReMBERT: 0.819 vs.\ 0.786 for Russian,
0.883 vs.\ 0.863 for Kazakh), showing that the
synthetic-trained detectors retain high classification performance on
this real-world split. The same pattern holds on ASR
transcripts: with encoders fine-tuned on transcribed text, real-world
macro-F1 remains high (.753--.835), and the O$\rightarrow$T degradation
grows with per-language ASR error ($\leq$6 points for Russian---absent
for ReMBERT, whose transcript fine-tuning even compensates for ASR
noise---and up to 12.5 points for Kazakh), mirroring the RQ1 trend. On the 290 parallel Russian--Kazakh fake pairs,
mDeBERTa's predictions disagree on only 3.8\% of items, indicating
substantial cross-lingual stability on identical false content.}

\rev{\paragraph{Provenance-controlled analysis.} 
\rev{We conduct this analysis on Russian original text.} In the synthetic binary
task, provenance (human vs.\ LLM) is perfectly correlated with the
label; a detector could therefore succeed by recognizing
machine-generated style alone. We complete the provenance$\times$veracity
matrix (Table~\ref{tab:confound}) with the real-world fakes
(human-written false) and with 290 \emph{faithful} LLM rewrites of the
same truthful articles, generated with Gemini~2.0 Flash-Lite under a
hallucination-free instruction. The results reveal evidence of two components. First, a veracity component: within
human-written text, both encoders separate false from truthful articles
by a wide margin (flag rates 1.000 vs.\ 0.352 for ReMBERT; 0.962 vs.\
0.221 for mDeBERTa). Second, a provenance component whose strength is
model-dependent: rewriting a truthful article---changing style while
preserving every fact---raises ReMBERT's flag rate from 0.352 to 1.000
(saturation on machine-generated text), but mDeBERTa's only to 0.586,
which remains well below its 0.893 rate on synthetic
hallucinations---i.e., mDeBERTa retains veracity sensitivity even within
machine-generated text. Synthetic binary F1 therefore overstates
factuality sensitivity, the degree of provenance reliance is itself a
model property, and the real-world split—where both classes originate from human-written content—substantially reduces the original provenance confound.}

\begin{table}[t]
\caption{\rev{Binary detection (macro-F1) on the synthetic test split
vs.\ the real-world split, for original text (O) and ASR transcripts
(T). Encoders are trained on the synthetic subset only, in a dedicated
run whose train/dev pool excludes the 290 truthful negatives per
language (Appendix~\ref{app:training}); the synthetic test split is
augmented with these negatives, so the truthful class of the Synth
(test) and Real rows shares the same 290 articles per language. O and T
columns use encoders fine-tuned on original text and on ASR transcripts,
respectively, over identical train/dev/test ids. These figures stem from
an independent fine-tuning run and are not directly comparable to
Table~\ref{tab:encoder_performance}, which reports the full corpus
split.}}
\label{tab:realworld}
\centering
\footnotesize
\renewcommand{\arraystretch}{1.02}
\setlength{\tabcolsep}{4pt}
\rev{
\begin{tabular}{llcccc}
\toprule
 & & \multicolumn{2}{c}{\textbf{ReMBERT}} & \multicolumn{2}{c}{\textbf{mDeBERTa}} \\
\cmidrule(lr){3-4}\cmidrule(lr){5-6}
\textbf{Lang} & \textbf{Split} & \textbf{O} & \textbf{T} & \textbf{O} & \textbf{T} \\
\midrule
\multirow{2}{*}{ru}
 & Synth (test) & .786 & .819 & .825 & .771 \\
 & Real         & .819 & .835 & .870 & .811 \\
\midrule
\multirow{2}{*}{kz}
 & Synth (test) & .863 & .794 & .831 & .748 \\
 & Real         & .883 & .818 & .878 & .753 \\
\bottomrule
\end{tabular}
}
\end{table}

\begin{table}[t]
\caption{\rev{Provenance$\times$veracity analysis: fraction of items
predicted ``hallucinated'' by ReMBERT (Russian, original text).
Human/Truthful = truthful negatives of the real-world split
(Section~\ref{ssec:realworld}); Human/False = factcheck.kz
fakes; LLM/Truthful = faithful Gemini~2.0 Flash-Lite rewrites of the same
truthful articles; LLM/False = synthetic hallucinations (test split).
mDeBERTa shows the same pattern with a weaker provenance component
(human-written .221/.962; LLM-generated .586/.893).}}
\label{tab:confound}
\centering
\footnotesize
\renewcommand{\arraystretch}{1.02}
\setlength{\tabcolsep}{6pt}
\rev{
\begin{tabular}{lcc}
\toprule
 & \textbf{Truthful} & \textbf{False} \\
\midrule
Human-written & .352 & 1.000 \\
LLM-generated & 1.000 & .986 \\
\bottomrule
\end{tabular}
}
\end{table}

\section{Insights and Challenges}
\label{sec:insights}

\textbf{1) Transcription quality is a key factor.}
Fine-tuned encoders generally degrade on ASR transcripts ($D_d$) relative to
original text ($D_a$). For mDeBERTa and ReMBERT this binary-task degradation
grows with per-language ASR error, largest in Kazakh, but the trend is less
consistent for fine-grained tasks.

\textbf{2) Zero-shot decoders mostly fail, but not uniformly.}

\new{Of five decoders (1.5B--8B), only Gemma-3n exceeds the majority-class
baseline on the binary task in macro-F1, on transcripts
(.613/.581/.630 vs.\ .40--.42; its audio responses were not retained).
The other four do not exceed their respective binary
macro-F1 baselines on transcripts, though Qwen2-Audio
exceeds the fine-grained baselines
(.198--.357 vs.\ .167--.176).
Several models evaluated outside their documented language coverage produce
near-constant outputs, so zero-shot multilingual audio reasoning is limited
first by language coverage and output reliability, and only then by task
granularity.}

\rev{\textbf{3) Low-resource conditions amplify pipeline challenges.}
\new{For the fine-tuned encoders,} Kazakh has the highest ASR error and
pronounced downstream degradation, consistent with compounded
TTS$\rightarrow$ASR noise.}

\textbf{4) Granularity increases difficulty.}

\rev{Binary classification is substantially easier than predicting alteration
type and severity; the low fine-grained scores show that these
construction-defined categories remain hard to distinguish reference-free.}

\textbf{5) Direct audio and transcripts are not comparable here.}
Evaluation coverage and class balance differ across audio and
transcript conditions. Audio items are long (median 96\,s), but
we did not verify whether inputs were truncated during inference;
any resulting information loss remains a possible,
model-dependent limitation. We therefore report audio results
without directly comparing them to transcript results.

\rev{\textbf{6) The binary signal reflects both veracity and provenance.}}
Detectors trained on synthetic hallucinations retain high performance on the
real-world split, with real-world F1 matching or exceeding synthetic-test F1
for both encoders and languages. However, \rev{in the Russian original-text
analysis,} faithful LLM rewrites of truthful articles reveal a
model-dependent provenance component: \new{rewriting a truthful article
without altering any fact raises ReMBERT's flag rate from .352 to 1.000, so
its binary signal is in large part machine-text detection, whereas mDeBERTa's
rises only from .221 to .586, well below its .893 on synthetic
hallucinations.} Synthetic binary scores thus overstate factuality
sensitivity, and provenance-controlled analyses are needed to quantify this
reliance, although other source-, register-, and distribution-related
confounds may remain.

\rev{Overall, the benchmark studies reference-free recognition of controlled,
text-level alterations rendered via TTS, not spontaneous speech or
evidence-grounded verification.} \rev{This controlled setup enables
systematic analysis of speech-pipeline and resource effects on text-based
detection, while exposing current limitations of zero-shot audio-language
models.} \rev{The real-world split partially relaxes this
constraint by grounding evaluation in naturally occurring misinformation.}

\section{Conclusion}
\label{sec:conclusion}

\rev{We introduce a multilingual spoken benchmark for reference-free
classification of synthetically introduced factual and contextual
alterations, comprising 12,013 aligned news samples in English, Russian,
and Kazakh. The resource represents source and rewritten articles as
text, synthesized speech, and ASR transcripts, and is complemented by a
real-world evaluation split of fact-checked misinformation in Russian
and Kazakh. It supports evaluation across text, transcript, and audio
without claiming evidence-grounded verification of factual truth.}

\rev{Results show that performance is strongly influenced by input
representation and language resources: for fine-tuned encoders, original
text generally outperforms ASR transcripts, with pronounced degradation
in low-resource Kazakh.} \new{In zero-shot settings, only Gemma-3n beat
the binary majority-class baseline (macro-F1) on transcripts (its audio
responses were not retained); the other four decoders did not exceed
their respective baselines.} \rev{On the
real-world split, synthetic-trained detectors
retain high classification performance, while the Russian provenance
analysis reveals both veracity-related and model-dependent machine-style
signals.}

We identify three main challenges: robustness to speech pipeline noise, amplified degradation in low-resource settings,
 and difficulty in fine-grained type and severity classification. Future work should focus on noise-robust multimodal modeling\rev{, provenance-robust detection objectives, and encoders that handle long-form audio}.

\section*{Limitations}

First, the benchmark primarily uses read speech synthesized with
language-specific TTS systems rather than spontaneous or natively recorded
speech. Its ASR transcripts consequently reflect errors introduced jointly
by synthesis and recognition, and the observed trends may not transfer
directly to natural-speech deployment. Human validation is limited to 120
Kazakh samples and does not cover all languages, voices, or experimental
conditions.

Second, WER is not directly comparable across typologically different
languages because it is sensitive to tokenization, segmentation, and
morphology. In agglutinative Kazakh, a minor morphological deviation can
produce a full word-level error. We therefore report CER alongside WER and
relate per-language ASR error to downstream performance rather than treating
raw WER as a direct measure of language difficulty.

Third, the benchmark covers only English, Russian, and Kazakh within the
news domain. Moreover, the English alterations were generated using GPT
models, whereas the Russian and Kazakh alterations were generated using
Gemini. Language is therefore partially confounded with generator identity,
and cross-lingual differences cannot be attributed exclusively to language
or resource availability. External validation by two cross-family LLM judges
and targeted human inspection reduce, but do not eliminate, annotation,
prompt, or generator-specific artifacts.

Fourth, provenance is correlated with the binary label in the synthetic
subset: original articles are human-written, whereas altered articles are
LLM-rewritten. A classifier may therefore respond partly to machine-generated
style rather than exclusively to veracity-related content. We examine this
confound in Section~\ref{ssec:rq3} using human-written misinformation and
faithful LLM rewrites. The type and severity tasks do not share this specific
binary confound, although they may retain prompt- or generator-specific cues.

Fifth, the evaluated models operate reference-free: they receive the target
text, transcript, or audio, but neither the source article nor retrieved
evidence. We also do not control whether individual news facts precede or
follow each model's training-data cutoff. The results should therefore be
interpreted as reference-free classification of synthetic alterations defined
relative to source articles, rather than evidence-grounded, open-world factual
verification.

Sixth, the real-world split contains 225 natively Russian and 65 natively
Kazakh misinformation items, translated into the other language to produce
290 items per language. The larger translated share of the Kazakh split may
introduce translation artifacts, and the split excludes English and is
intended only for evaluation. In addition, fact-checked misinformation is not
a random sample, and the real-world positives differ from the truthful
negatives in source and register as well as veracity. High performance on
this split may therefore partly reflect distributional cues. The
provenance-controlled analysis reduces the direct human-versus-LLM confound
but does not eliminate all such differences.

Finally, the decoder models are evaluated with zero-shot prompting and
without task-specific audio fine-tuning. \rev{None of them documents support for Kazakh, and the two Qwen models do
not report the languages their speech encoder covers,} so results on these
languages partly reflect model coverage rather than task difficulty. Their
results characterize out-of-the-box performance under the evaluated
conditions and should not be interpreted as an upper bound on the
usefulness of audio. Fine-tuned or evidence-grounded multimodal models
may exhibit different modality patterns.

\section*{Ethical Considerations}

The benchmark is built from news articles published by public Kazakhstani
outlets (nur.kz, forbes.kz, tengrinews.kz, kapital.kz). To respect
third-party copyright, we do not redistribute the original article texts
or their verbatim derivatives. For original articles and real-world fake
items, the release contains only source URLs, labels, and metadata; ASR
transcripts and synthesized audio are released only for LLM-rewritten
items and can be regenerated for the original texts with the publicly
available TTS and ASR systems described in Section~\ref{ssec:speech}. The
release also includes our LLM-generated hallucinated rewrites,
type/severity labels, and the split manifests.

\rev{For the small fraction of articles whose original pages are no longer
available online, we provide the source outlet and publication metadata in
place of a URL.}

The original article texts are used internally solely as factual
references. Because the rewrites necessarily retain part of the source
content, their redistribution is subject to the limitations set out in
Article~19 of the Republic of Kazakhstan Law on Copyright and Related
Rights (No.~6-I), including attribution of each item to its source and use
limited to the extent justified by the research purpose. The corpus
consists only of already-published news and contains no private or
sensitive personal data.

\rev{The real-world split is built from misinformation items publicly
identified and debunked by factcheck.kz, a professional Kazakhstani
fact-checking organization. We do not redistribute the fake texts, their
translations, or their ASR transcripts and synthesized audio; the release
contains only URLs of the published fact-checking reports, binary labels,
and translation metadata, following the same scheme as for the original
articles. Because every item in this split has already been publicly
debunked and no fake text is redistributed, our resource does not
introduce new misinformation into circulation.}

Because the resource contains deliberately fabricated and contradictory
content, it carries a potential for misuse (e.g., as disinformation).
Rewrites may also attribute fabricated statements to real, named
individuals and organizations; every rewritten item is therefore labeled
as a synthetic alteration, and the dataset card states that rewrites must
not be presented as factual reporting. The resource is intended solely for developing and evaluating detection
systems and is distributed under a custom research-use license
(\texttt{lost-in-speech-research-use-license}; full text in the
\texttt{LICENSE} file of the dataset repository).

We used generative models (\texttt{gpt-3.5-turbo}, \texttt{gpt-4}, and
\texttt{gemini-2.0-flash-lite}) to generate hallucinated text and
\texttt{claude-haiku-4-5-20251001} and \texttt{deepseek-chat} as independent
LLM-based judges for partial quality evaluation; this use is disclosed here
in accordance with ACL policy on the use of generative AI.

\section*{Acknowledgments}

This work was supported in part by U.S. NSF Awards
\#2438810 and \#2555559, an Amazon Nova AI Challenge Award,
and the Penn State Frymoyer Chairship. Some experimental
results were obtained using computational resources provided
by CloudBank, supported through U.S. NAIRR Award \#240336.

\bibliography{custom}

\appendix

\newenvironment{promptbox}[1]{
  \par\medskip\noindent\hrule\nobreak\smallskip\nobreak
  \noindent{\small\bfseries #1}\par\nobreak\smallskip\nobreak

  \footnotesize\fontencoding{T2A}\fontfamily{cmtt}\selectfont\frenchspacing
  \setlength{\parindent}{0pt}\setlength{\parskip}{2pt}\obeylines}
  {\par\nobreak\smallskip\hrule\medskip}

\section{\rev{Generation and Evaluation Prompts}}
\label{app:prompts}

This appendix documents, verbatim, the prompt templates of the final
pipeline. Runtime variables are replaced by braced placeholders in
capital letters (Table~\ref{tab:placeholders}); all other characters,
including typographical irregularities, are reproduced exactly as sent
to the models and marked in footnotes rather than corrected. For each stage we show one canonical template in full; language variants
are reported as verbatim line-level deltas. Superseded pilot prompts and
auxiliary preprocessing prompts, which did not contribute to the released
corpus, are omitted.\footnote{The dataset
(\url{https://huggingface.co/datasets/maristombayeva/lost-in-speech})
includes the LLM-rewritten texts, type/severity and binary labels,
source references, ASR transcripts and synthetic audio for the rewritten
items in all three languages, the split manifests of both training runs,
and encoder predictions. The dataset card documents record-level
exceptions and access conditions.}
\paragraph{Generation conditions.}
\begin{itemize}
  \item \textbf{Generation:} \texttt{gpt-3.5-turbo} and \texttt{gpt-4}
        for the English batches; \texttt{gemini-2.0-flash-lite} for the
        Russian and Kazakh batches. For \texttt{gemini-2.0-flash-lite},
        default API decoding parameters (no temperature or top-$p$
        overrides set).
  \item \textbf{Judge:} \texttt{claude-haiku-4-5-20251001} (temperature $0$,
        \texttt{max\_tokens} $300$) and \texttt{deepseek-chat}
        (temperature $0$); identical prompt for both.
    \item \textbf{Audio/text detection:} all five decoders
        (Table~\ref{tab:model_support}) use greedy decoding with
        \texttt{max\_new\_tokens} $256$.
\end{itemize}

\begin{table*}[t]
\centering\small
\resizebox{\textwidth}{!}{
\begin{tabular}{ll}
\toprule
Placeholder & Runtime value \\
\midrule
\{ARTICLE TEXT\} & Source news article passed to the generation model. \\
\{SEVERITY LEVEL DESCRIPTION\} & Full mild/moderate/severe definition string (severity-definitions box). \\
\{HALLUCINATION TYPE\} & \texttt{contradiction} / \texttt{fabrication} / \texttt{context\_inconsistency}. \\
\{SEVERITY LEVEL\} & \texttt{mild} / \texttt{moderate} / \texttt{severe}. \\
\{SEVERITY LEVEL DEFINITION\} & Judge-side one-sentence definition of the (type, level) pair. \\
\{DIMENSION LIST\} / \{DIMENSION KEYS\} & Type-dependent dimension names (list / quoted JSON keys). \\
\{ORIGINAL ARTICLE TEXT\}, \{REWRITTEN ARTICLE TEXT\} & Article pair for the judge (each truncated to $6{,}000$ chars). \\
\{INPUT TEXT\} & Text passage under detection. \\
\{DIGITS RULE\} & \texttt{Remove all digits.} or \texttt{Keep digits.} \\
\bottomrule
\end{tabular}
}
\caption{Placeholders used in the prompt templates.}
\label{tab:placeholders}
\end{table*}

\subsection{Hallucination Generation}

\begin{promptbox}{Prompt for factual fabrication generation (canonical, English)}
You are a professional news editor tasked with generating factual fabrications at varying levels. Here is the original news article:
~
Original news: \{ARTICLE TEXT\}
~
Follow this structure:
1. Extract the narrative, factual information, entities, and contextual data from the original news.
2. Create one version of the news:
~~~- \{SEVERITY LEVEL DESCRIPTION\}
~
3. Provide a report:
~~~- Factual accuracy
~~~- Credibility
~~~- Fluency
~~~- Coherence
~~~- Relevance
~~~- Satisfaction with fabrication level
~~~- Adequacy of type
~
~~~Evaluate each dimension as: Low, Very Low, Medium, High, Very High.
~
4. Identify the fabricated sentences and explain the deviations.
~
5. Make sure to provide the **full text of the news with fabrications**, without hyperlinks:
~~~- \{SEVERITY LEVEL DESCRIPTION\} (full text):
~
Generate the answer strictly in English and strictly following this format.
\end{promptbox}

Instantiated once per article and severity level; the full news text is
extracted by matching the ``(full text)'' header.

\begin{promptbox}{Prompt for factual contradiction generation (canonical, English)}
You are a professional news editor tasked with generating factual contradictions at varying levels. Here is the original news article:
~
Original news: \{ARTICLE TEXT\}
~
Follow this structure:
1. Extract the narrative, factual information, entities, and contextual data from the original news.
2. Create one version of the news focusing on factual contradictions:
~~~~- \{SEVERITY LEVEL DESCRIPTION\}
~
3. Provide a report:
~~~~- Factual accuracy
~~~~- Credibility
~~~~- Fluency
~~~~- Coherence
~~~~- Relevance
~~~~- Satisfaction with contradiction level
~~~~- Adequacy of type
~
~~~~Evaluate each dimension as: Low, Very Low, Medium, High, Very High.
~
4. Identify the sentences containing factual contradictions and explain the deviations from the original text and real-world facts.
~
5. Make sure to provide the **full text of the news with factual contradictions**, without hyperlinks:
~~~~- \{SEVERITY LEVEL DESCRIPTION\} (full text):
~
Generate the answer strictly in English and strictly following this format.
\end{promptbox}

\begin{promptbox}{Language-variant deltas: fabrication and contradiction (RU/KZ)}
The RU/\allowbreak{}KZ variants differ from the canonical templates ONLY in the following verbatim lines.
~
[Step 1] "...from the original news" is extended with the target language:
1. Extract the narrative, factual information, entities, and contextual data from the original news in **Russian**.
1. Extract the narrative, factual information, entities, and contextual data from the original news in **Kazakh**.
~
[Step 2] An additional bullet is appended after \{SEVERITY LEVEL DESCRIPTION\}:
~~~~- **Ensure the length of the generated news text is similar to the original. Avoid shortening or excessively extending the article. Maintain paragraph structure and detail level.**
~
[Step 3] Header and scale are localized (dimension names keep trailing colons):
3. Provide a report evaluation in **Russian**:
~~~~Evaluate each dimension as: Низкая, Очень низкая, Средняя, Высокая, Очень высокая.
3. Provide a report evaluation in **Kazakh**:
KZ fabrication scale:
~~~~Evaluate each dimension as: Өте төмен, Төмен, Орташа, Жоғары, Өте жоғары
KZ contradiction scale:
~~~~Evaluate each dimension as: Өте төмен, Төмен, Орташа, Жоғары, Өте жоғары.
~
[Step 4] "...explain the deviations in **Russian**." /\allowbreak{} "...real-world facts in **Russian**." (same pattern with **Kazakh**)
~
[Step 5] Output-language requirement and localized full-text markers:
5. Make sure to provide the **full text of the news with fabrications** in **Russian**, without hyperlinks:
~~~~- \{SEVERITY LEVEL DESCRIPTION\} (полный текст):
KZ fabrication marker:
~~~~- \{SEVERITY LEVEL DESCRIPTION\} (толық мәтін):
KZ contradiction marker:
~~~~- \{SEVERITY LEVEL DESCRIPTION\} (толық мәтін):
~
[Closing line]
Generate the answer strictly in English for the instructions and strictly following this format. The report evaluation and the full text with fabrications should be in **Russian**.
Generate the answer strictly in English for the instructions and strictly following this format. The report evaluation and the full text with fabrications should be in **Kazakh**.
Generate the answer strictly in English for the instructions and strictly following this format. The report evaluation and the full text with factual contradictions should be in **Russian**.
Generate the answer strictly in English for the instructions and strictly following this format. The report evaluation and the full text with factual contradictions should be in **Kazakh**.
\end{promptbox}

\paragraph{Translation and transliteration note.}
The non-English strings appearing above are translated and transliterated
as follows. Russian
``\cyrtext{Низкая, Очень низкая, Средняя, Высокая, Очень высокая}''
(\textit{Nizkaya, Ochen' nizkaya, Srednyaya, Vysokaya,
Ochen' vysokaya}) means ``Low, Very Low, Medium, High, Very High''.
Kazakh
``\cyrtext{Өте төмен, Төмен, Орташа, Жоғары, Өте жоғары}''
(\textit{Öte tömen, Tömen, Ortasha, Zhogary, Öte zhogary})
means ``Very Low, Low, Medium, High, Very High''.
Russian ``\cyrtext{полный текст}'' (\textit{polnyy tekst})
and Kazakh ``\cyrtext{толық мәтін}'' (\textit{tolyq mätin})
both mean ``full text''.

\begin{promptbox}{Prompt for context inconsistency generation (canonical, English)}
You are a professional news editor tasked with generating a rewritten news article that introduces **context inconsistency** at a specified level. You will be provided with an original article.
~
Your goal is to alter the logical or narrative structure (entities, causality, timeline, etc.) while keeping all **explicit numeric facts** untouched.
~
Original news article:
\{ARTICLE TEXT\}
~
Your task:
~
1. Carefully extract and preserve all **explicit numeric values** from the original article. This includes:
~~~- Years and dates (e.g., 2023, July 15)
~~~- Quantities (e.g., 5,000 employees)
~~~- Money amounts (e.g., \$1.2 billion)
~~~- Percentages (e.g., 13.7\%)
~~~- Measurable data (e.g., 75 km, 3.6 GHz)
~
2. Create a **rewritten version** of the article that demonstrates **context inconsistency** as described below:
~~~~- \{SEVERITY LEVEL DESCRIPTION\}
~
~~~~Critical rules:
~~~~- DO NOT add, remove, change, or contradict ANY numeric values from the original article. They must appear exactly as in the original.
~~~~- All inconsistencies must arise from contradictions or distortions in logic, narrative flow, cause-effect chains, timeline, or facts — but NEVER from numerical manipulation.
~~~~- Output length must be within ±15\% of the original article length.
~
3. Provide a quality report evaluating the rewritten article:
~~~~- Internal Consistency
~~~~- Logical Coherence
~~~~- Fluency
~~~~- Overall Coherence
~~~~- Relevance to Original News
~~~~- Satisfaction with Inconsistency Level
~~~~- Adequacy of Inconsistency Type
~~~~Rate each using: Very Low, Low, Medium, High, Very High.
~
4. Identify and list all sentences that introduce contradictions, context misinterpretations, or logical flaws. Explain why they are inconsistent with the original article.
~
5. Output the full rewritten news article with the specified inconsistency level:
~
\{SEVERITY LEVEL DESCRIPTION\} (full text):
\end{promptbox}

Unlike the other two types, this template forbids numeric changes, adds a
$\pm15\%$ length constraint, and rates Internal Consistency and Logical
Coherence instead of Factual accuracy and Credibility.

\begin{promptbox}{Language-variant deltas: context inconsistency (RU/KZ)}
The RU/\allowbreak{}KZ variants differ from the canonical template ONLY in the following verbatim lines.
~
[Intro] "...an original article in Russian." /\allowbreak{} "...an original article in Kazakh."
Original news article (in Russian):
Original news article (in Kazakh):
~
[Step 1] Localized numeric examples:
~~~~- Years and dates (e.g., 2023, 15 июля)
~~~~- Quantities (e.g., 5 000 сотрудников)
~~~~- Money amounts (e.g., \$1.2 миллиарда)
~~~~- Measurable data (e.g., 75 км, 3.6 ГГц)
~~~~- Years and dates (e.g., 2023, 15 шілде)
~~~~- Quantities (e.g., 5000 қызметкер)
~~~~- Money amounts (e.g., \$1,2 млрд)
~
[Step 2]
2. Create a **rewritten version** of the article (in Russian) that demonstrates **context inconsistency** as described below:
(same pattern with "(in Kazakh)")
~
[Step 3] Dimension names keep trailing colons; localized scales:
~~~~Rate each using: Очень низкий, Низкий, Средний, Высокий, Очень высокий.
~~~~Rate each using: Өте төмен, Төмен, Орташа, Жоғары, Өте жоғары.
~
[Step 4]
4. Identify and list all sentences (in Russian) that introduce contradictions, context misinterpretations, or logical flaws. Explain (in Russian) why they are inconsistent with the original article.
(same pattern with "(in Kazakh)")
~
[Step 5]
5. Output the full rewritten news article with the specified inconsistency level (in Russian):
\{SEVERITY LEVEL DESCRIPTION\} (полный текст на русском языке):
5. Output the full rewritten news article with the specified inconsistency level (in Kazakh):
\{SEVERITY LEVEL DESCRIPTION\} (толық мәтіні қазақ тілінде):
\end{promptbox}

\paragraph{Translation and transliteration note.}
The additional non-English examples above are translated and transliterated
as follows. Russian ``\cyrtext{15 июля}''
(\textit{15 iyulya}), ``\cyrtext{5 000 сотрудников}''
(\textit{5 000 sotrudnikov}), ``\cyrtext{\$1.2 миллиарда}''
(\textit{\$1.2 milliarda}), and ``\cyrtext{75 км, 3.6 ГГц}''
(\textit{75 km, 3.6 GGts}) mean ``15 July'', ``5,000 employees'',
``\$1.2 billion'', and ``75 km, 3.6 GHz'', respectively.
Kazakh ``\cyrtext{15 шілде}''
(\textit{15 shilde}), ``\cyrtext{5000 қызметкер}''
(\textit{5000 qyzmetker}), and ``\cyrtext{\$1,2 млрд}''
(\textit{\$1,2 mlrd}) mean ``15 July'', ``5,000 employees'',
and ``\$1.2 billion'', respectively.
Russian ``\cyrtext{полный текст на русском языке}''
(\textit{polnyy tekst na russkom yazyke}) means ``full text in Russian'',
and Kazakh ``\cyrtext{толық мәтіні қазақ тілінде}''
(\textit{tolyq mätini qazaq tilinde}) means ``full text in Kazakh''.
The localized rating scales use the same translations and transliterations
as reported above.

\begin{promptbox}{Severity-level definitions inserted as \{SEVERITY LEVEL DESCRIPTION\}}
Factual fabrication:
- Mild fabrication: Introduce small factual deviations (e.g., slightly incorrect dates, locations, or names) that are still plausible and do not alter the overall message of the article.
- Moderate fabrication: Introduce multiple fictional or unverifiable statements that are plausible but subtly distort the historical or factual context of the article.
- Severe fabrication: Rewrite the article by introducing clearly false or unverifiable claims that significantly contradict established facts or distort the original narrative.
~
Factual fabrication — Moderate/\allowbreak{}Severe variants used for the RU/\allowbreak{}KZ Gemini batches:
- Moderate fabrication: Introduce multiple fictional or unverifiable statements that are plausible and subtly distort the historical or factual context of the article. Ensure the narrative remains within the realm of possibility.
- Severe fabrication: Rewrite the article by introducing significant factual errors and unverifiable claims that substantially alter the narrative. However, maintain a degree of realism and avoid overtly fantastical or scientifically impossible elements. Focus on plausible alternative scenarios or large-scale misinformation.
~
Factual contradiction (all languages):
- Mild factual contradiction: Introduce minor factual errors involving entities or relations that directly contradict easily verifiable information but maintain the overall narrative structure.
- Moderate factual contradiction: Introduce several factual errors involving key entities and their relationships, significantly distorting the factual basis of the news while maintaining a degree of plausibility in the narrative.
- Severe factual contradiction: Completely rewrite significant portions of the news with demonstrably false and contradictory information regarding core entities, events, and their relationships, resulting in a highly inaccurate account.
~
Context inconsistency (all languages):
- Mild context inconsistency: Introduce minor contradictions that deviate from specific details or assumptions in the original article, while largely preserving its narrative flow.
- Moderate context inconsistency: Introduce several contradictions that directly oppose key elements or implications of the original article, potentially creating some context inconsistencies.
- Severe context inconsistency: Generate a version of the article that fundamentally disregards and contradicts the original article’s internal context, introducing major logical flaws and factual distortions.
\end{promptbox}

The fabrication definitions were revised between the English batch and
the later Russian/Kazakh batches; both wordings contributed articles to
the corpus and are therefore reported.

\subsection{LLM-as-a-Judge Quality Evaluation}

\begin{promptbox}{Prompt for LLM-as-a-judge quality evaluation (Claude and DeepSeek)}
You are a professional news-quality evaluator (judge only; do NOT rewrite the article).
You are given the ORIGINAL article, the intended hallucination TYPE and SEVERITY, and a REWRITTEN article.
Rate the REWRITTEN article on each dimension using EXACTLY one of:
Very Low, Low, Medium, High, Very High.
~
TYPE: \{HALLUCINATION TYPE\}
SEVERITY: \{SEVERITY LEVEL\} — \{SEVERITY LEVEL DEFINITION\}
~
Dimensions:
\{DIMENSION LIST\}
~
Return ONLY a JSON object mapping each dimension name to its rating, using these exact keys:
\{\{DIMENSION KEYS\}\}
~
ORIGINAL:
\{ORIGINAL ARTICLE TEXT\}
~
REWRITTEN:
\{REWRITTEN ARTICLE TEXT\}
\end{promptbox}

\begin{promptbox}{Type-dependent dimension sets inserted as \{DIMENSION LIST\} / \{DIMENSION KEYS\}}
contradiction:  Factual accuracy (FA), Credibility (Cr), Fluency (Fl), Coherence (Co), Relevance (Re), Satisfaction with contradiction level (Sa), Adequacy of type (AT)
~
fabrication:    Factual accuracy (FA), Credibility (Cr), Fluency (Fl), Coherence (Co), Relevance (Re), Satisfaction with fabrication level (Sa), Adequacy of type (AT)
~
context\_inconsistency:  Internal Consistency (IC), Logical Coherence (LC), Fluency (Fl), Overall Coherence (Co), Relevance to Original News (Re), Satisfaction with Inconsistency Level (Sa), Adequacy of Inconsistency Type (AT)
\end{promptbox}

\begin{promptbox}{Judge-side severity definitions inserted as \{SEVERITY LEVEL DEFINITION\}}
contradiction:
- mild: Mild factual contradiction: minor factual errors involving entities or relations that directly contradict easily verifiable information but maintain the overall narrative structure.
- moderate: Moderate factual contradiction: several factual errors involving key entities and relationships, significantly distorting the factual basis while keeping some plausibility.
- severe: Severe factual contradiction: rewrite significant portions with demonstrably false, contradictory information about core entities, events, and relationships.
~
fabrication:
- mild: Mild fabrication: small plausible factual deviations (e.g., slightly incorrect dates, locations, names) that do not alter the overall message.
- moderate: Moderate fabrication: multiple fictional or unverifiable but plausible statements that subtly distort the factual context.
- severe: Severe fabrication: clearly false, unverifiable, or fantastical claims that significantly contradict real-world knowledge.
~
context\_inconsistency:
- mild: Mild context inconsistency: subtle shifts in emphasis or context that slightly distort meaning without explicit factual errors.
- moderate: Moderate context inconsistency: noticeable distortions of context/\allowbreak{}emphasis that change interpretation.
- severe: Severe context inconsistency: strong contextual distortions that substantially change the meaning of the article.
\end{promptbox}

Both judges belong to model families disjoint from the generators.
Verbal ratings are mapped to integers $1$--$5$ during parsing (including
Kazakh scale variants); the JSON object is extracted with a regular
expression, with up to three retries on parse failure.

\subsection{Zero-Shot Detection (All Decoders, Audio and Text)}

\begin{promptbox}{System prompt for Qwen2.5-Omni}
You are Qwen, a virtual human developed by the Qwen Team, Alibaba Group, capable of perceiving auditory and visual inputs, as well as generating text and speech.
\end{promptbox}

All five decoders receive the same user-turn prompts below. The system
prompt above is the default system prompt of Qwen2.5-Omni; the other
decoders use their default system prompt.

\begin{promptbox}{Prompt for direct zero-shot detection, binary task (canonical, audio input)}
You are an expert assistant specialized in analyzing audio content and detecting hallucinations.
~
Your task is to determine whether there are any hallucinations in the audio content.
~
A hallucination is any statement that contains factual contradictions, fabricated details, or contextual inconsistencies relative to the actual audio content.
~
Given the audio content, determine:
Is there a hallucination? (yes/\allowbreak{}no)
~
Output format: \{"binary": "yes/\allowbreak{}no"\}
\end{promptbox}

\begin{promptbox}{Type and severity task variants (verbatim components)}
The type and severity prompts follow the same structure as the binary prompt, with the following verbatim substitutions.
~
TYPE task — role, task statement, definitions, and closing:
You are an expert assistant specialized in analyzing audio content and classifying hallucination types.
Your task is to identify the type of hallucination present in the audio content.
~
Hallucination Types:
- Factual Contradiction: Statements that directly conflict with known facts or information provided in the audio content.
- Factual Fabrication: Insertion of fabricated yet plausible-sounding details not grounded in the audio content.
- Contextual Inconsistency: Subtle alterations that distort the meaning, emphasis, or context of the audio content without introducing explicit factual errors.
~
Given the audio content, classify the hallucination type:
What type of hallucination is present? (factual\_contradiction/\allowbreak{}factual\_fabrication/\allowbreak{}contextual\_inconsistency/\allowbreak{}none)
~
Output format: \{"type": "factual\_contradiction/\allowbreak{}factual\_fabrication/\allowbreak{}contextual\_inconsistency/\allowbreak{}none"\}
~
SEVERITY task — role, task statement, definitions, and closing:
You are an expert assistant specialized in analyzing audio content and assessing hallucination severity.
Your task is to assess the severity level of any hallucination present in the audio content.
~
Severity Levels:
- Mild: Subtle distortions or minor deviations that preserve the main narrative and plausibility of the audio content.
- Moderate: Noticeable inconsistencies or factual alterations that affect key details or context while maintaining partial alignment with the audio content.
- Severe: Major contradictions, fabrications, or contextual breakdowns that substantially misrepresent or conflict with the audio content's facts or intent.
~
Given the audio content, classify the hallucination severity:
What is the severity level? (mild/\allowbreak{}moderate/\allowbreak{}severe/\allowbreak{}none)
~
Output format: \{"degree": "mild/\allowbreak{}moderate/\allowbreak{}severe/\allowbreak{}none"\}
\end{promptbox}

\begin{promptbox}{Chain-of-thought variants (verbatim replaced closing lines)}
In the CoT condition the direct closing lines are replaced as follows (Output format lines unchanged).
~
BINARY:
Given the audio content, think step-by-step about whether hallucinations are present:
~
Then provide your final answer:
Is there a hallucination? (yes/\allowbreak{}no)
~
TYPE:
Given the audio content, think step-by-step about what type of hallucination may be present:
~
Then provide your final classification:
What type of hallucination is present? (factual\_contradiction/\allowbreak{}factual\_fabrication/\allowbreak{}contextual\_inconsistency/\allowbreak{}none)
~
SEVERITY:
Given the audio content, think step-by-step about the severity of any hallucination:
~
Then provide your final assessment:
What is the severity level? (mild/\allowbreak{}moderate/\allowbreak{}severe/\allowbreak{}none)
\end{promptbox}

Each task is a separate call under both conditions. Text-input variants
replace ``audio content'' with ``text content''; in the text condition
the passage is appended as ``\texttt{Text to analyze: \{INPUT TEXT\}}'',
while in the audio condition the audio file is attached to the user
turn. Predictions are parsed from the JSON fragment of the response.

\subsection{ASR Transcript Cleaning for WER/CER}

\begin{promptbox}{Prompt for ASR transcript cleaning for WER/CER (canonical, single item)}
Clean ASR transcript for WER/\allowbreak{}CER evaluation.
~
Rules:
- Do NOT add, paraphrase, or translate.
- Lowercase.
- Remove punctuation, quotes, brackets, noise tokens ([music], <unk>), timestamps (e.g., 00:01:23).
- Replace hyphens with spaces.
- Collapse multiple spaces.
- \{DIGITS RULE\}
~
Return ONLY cleaned text.
\end{promptbox}

\begin{promptbox}{Batched (JSONL) cleaning variant (verbatim deltas)}
The batched variant differs in the following verbatim lines:
~
You clean ASR transcripts for fair WER/\allowbreak{}CER evaluation.
- Do NOT add new content, do NOT paraphrase, do NOT translate.
- Remove punctuation, noise tokens and timestamps (e.g., 00:01:23).
~
OUTPUT FORMAT (very important):
Return ONLY JSONL: one JSON object per line, no surrounding array, no markdown.
Each line must be:
\{"idx": <int>, "cleaned": "<string>"\}
\end{promptbox}

Cleaning normalizes both reference and hypothesis strings before WER/CER
computation; batches are split recursively on JSON parse failures.

\section{\rev{Training Details}}
\label{app:training}
\rev{Encoders are fine-tuned for the binary task with effective batch
size 16 (batch 4, gradient accumulation 4), learning rate 2e-5, 3
epochs, maximum sequence length 512, fp16 with gradient checkpointing,
and early stopping on development macro-F1 (patience 1); seed 42. For
mDeBERTa-v3, which is unstable under fp16, we instead use fp32, eager
attention, learning rate 1e-5 with 10\% linear warmup, and physical
batch 8 with gradient accumulation 2 (same effective batch 16). Data
are split 80/10/10 (train/dev/test), stratified by language and label.
The encoders evaluated on the real-world split
(Table~\ref{tab:realworld}) are trained in a dedicated run in which the
580 truthful negatives (290 per language, drawn from the ru/kz
non-hallucinated class of the corpus) are removed from the train/dev
pool before splitting and appended to the synthetic test split, ensuring
that no model evaluated on real-world data was trained on its truthful
items. Table~\ref{tab:encoder_performance} reports an independent run of
the same configuration on the full corpus split; cross-table differences
in the binary figures therefore combine training-pool, test-composition,
and run-to-run effects, and within-table comparisons always use a single
run. The split files of both runs are included in the released dataset. For the transcript (T) columns of
Table~\ref{tab:realworld}, both encoders are fine-tuned on the ASR
transcripts of the synthetic corpus using the identical train/dev/test
split (by article id) as the original-text models.}

\section{\rev{Models}}
\label{app:models}

\rev{Table~\ref{tab:model_support} lists all detection, generation, and
speech models used in the study, with their modality support.
Table~\ref{tab:asr_choice} details the per-language ASR selection
discussed in Section~\ref{ssec:speech}.}

\begin{table}[t]
\caption{Per-language ASR error (WER/CER, \%, corpus level) on identical synthesized audio: a single multilingual model (Whisper-large-v3) vs. our per-language selection. Whisper-large-v3 fails on low-resource Kazakh, while a fine-tuned wav2vec2 model nearly halves the Kazakh error rate; we therefore use Whisper-large-v3 for English/Russian and fine-tuned wav2vec2 for Kazakh.}
\label{tab:asr_choice}
\centering
\footnotesize
\renewcommand{\arraystretch}{1.02}
\setlength{\tabcolsep}{3pt}
\resizebox{\columnwidth}{!}{
\begin{tabular}{l ccc}
\toprule
\textbf{ASR system} & \textbf{EN} & \textbf{RU} & \textbf{KZ} \\
 & WER/CER & WER/CER & WER/CER \\
\midrule
Whisper-large-v3        & \textbf{7.30/2.74} & \textbf{9.08/4.74} & 64.27/32.03 \\
wav2vec2 (fine-tuned)   & --                 & --                 & \textbf{34.04/16.90} \\
\midrule
\textit{Used in benchmark} & Whisper & Whisper & wav2vec2 \\
\bottomrule
\end{tabular}
}
\end{table}

\begin{table}[t]
\caption{\rev{Models used in the study.
P = Purpose ($\circ$ Detection, $\square$ Text generation, $\triangle$ Audio/TTS/ASR).
Type: Enc = Encoder, Dec = Decoder.
A = Audio support, T = Text support.
\textbf{Lang} = languages with officially documented support, per the model
cards or technical reports; for audio decoders the figure before the slash
refers to text and the figure after it to speech input (n/r = not reported
by the developers). TTS/ASR rows give supported languages, not the single
language each system is used for here (Section~\ref{ssec:speech}).
$^{\ddagger}$Gemma-3n-E4B has 8B raw parameters and runs with a memory
footprint comparable to a 4B model.}}
\label{tab:model_support}
\centering
\footnotesize
\renewcommand{\arraystretch}{1.05}
\setlength{\tabcolsep}{3pt}
\resizebox{\columnwidth}{!}{
\begin{tabular}{@{}l c c c c c c c@{}}
\toprule
\textbf{Model} & \textbf{P} & \textbf{Type} & \textbf{Lang} & \textbf{Params} & \textbf{Year} & \textbf{A} & \textbf{T} \\
\midrule
\multicolumn{8}{c}{\textit{Detection Models}} \\
\midrule
XLM-R (base)      & $\circ$ & Enc & 100 & 279M  & 2019 & \ding{55} & \ding{51} \\
XLM-R (large)     & $\circ$ & Enc & 100 & 561M  & 2019 & \ding{55} & \ding{51} \\
mDeBERTa          & $\circ$ & Enc & 100 & 279M  & 2021 & \ding{55} & \ding{51} \\
ReMBERT           & $\circ$ & Enc & 110 & 576M  & 2021 & \ding{55} & \ding{51} \\
Qwen2.5-Omni      & $\circ$ & Dec & 29\,/\,n/r & 3B & 2025 & \ding{51} & \ding{51} \\
Qwen2-Audio       & $\circ$ & Dec & 8\,/\,n/r  & 7B & 2024 & \ding{51} & \ding{51} \\
Gemma 3n          & $\circ$ & Dec & 140\,/\,35 & 4B$^{\ddagger}$ & 2025 & \ding{51} & \ding{51} \\
LFM2-Audio        & $\circ$ & Dec & en\,/\,en  & 1.5B & 2025 & \ding{51} & \ding{51} \\
Step-Audio-2-mini & $\circ$ & Dec & zh,en\,/\,zh,en & 8B & 2025 & \ding{51} & \ding{51} \\

\midrule
\multicolumn{8}{c}{\textit{Generation Models}} \\
\midrule
GPT-3.5-turbo     & $\square$ & Dec & 100+ & -- & 2023 & \ding{55} & \ding{51} \\
GPT-4             & $\square$ & Dec & 100+ & -- & 2023 & \ding{55} & \ding{51} \\
Gemini 2.0 Flash-Lite & $\square$ & Dec & 100+ & -- & 2025 & \ding{55} & \ding{51} \\
\midrule
Coqui XTTS-v2     & $\triangle$ & TTS & 17 (en used) & --   & 2023 & \ding{51} & \ding{55} \\
Silero TTS        & $\triangle$ & TTS & multi (ru used) & -- & 2021 & \ding{51} & \ding{55} \\
Kazakh TTS        & $\triangle$ & TTS & kz & 50M  & 2021 & \ding{51} & \ding{55} \\
\midrule
Whisper-large-v3  & $\triangle$ & ASR & 99 & 1550M & 2023 & \ding{51} & \ding{51} \\
wav2vec2 (fine-tuned) & $\triangle$ & ASR & kz & -- & 2024 & \ding{51} & \ding{51} \\
\bottomrule
\end{tabular}
}
\end{table}

\section{\rev{Full LLM-Based Quality Evaluation}}
\label{app:quality}

\rev{Table~\ref{tab:llm_eval} reports the complete LLM-based quality
evaluation by hallucination type, severity, and language, discussed in
Section~\ref{ssec:quality}. Table~\ref{tab:judge_agreement} reports
per-dimension agreement between the two judges.}

\begin{table}[t]
\caption{Cross-judge agreement on the stratified sample between the two
external judges, Claude Haiku~4.5 and DeepSeek Chat, per quality dimension.
$\kappa$: quadratic-weighted Cohen's $\kappa$; $\rho$: Spearman correlation;
W1: agreement within one point. Ordered by $\kappa$.}
\label{tab:judge_agreement}
\centering
\footnotesize
\renewcommand{\arraystretch}{1.02}
\setlength{\tabcolsep}{5pt}
\begin{tabular}{l r r r r}
\toprule
\textbf{Dimension} & $n$ & $\kappa$ & $\rho$ & \textbf{W1} \\
\midrule
Credibility (Cr)          & 144 & 0.897 & 0.923 & 1.00 \\
Factual Accuracy (FA)     & 144 & 0.875 & 0.910 & 1.00 \\
Relevance (Re)            & 216 & 0.676 & 0.721 & 0.92 \\
Satisfaction (Sa)         & 216 & 0.600 & 0.674 & 0.97 \\
Coherence (Co)            & 216 & 0.591 & 0.610 & 0.92 \\
Adequacy of Type (AT)     & 216 & 0.511 & 0.497 & 0.97 \\
Fluency (Fl)              & 216 & 0.311 & 0.390 & 0.92 \\
Logical Coherence (LC)    &  72 & 0.144 & 0.258 & 0.78 \\
Internal Consistency (IC) &  72 & 0.110 & 0.175 & 0.71 \\
\bottomrule
\end{tabular}
\end{table}

\begin{table}[t]
\caption{LLM-as-a-judge quality evaluation by hallucination type, severity,
and language, computed by the independent judge (claude-haiku-4-5,
temperature 0) on the stratified validation sample of Section~\ref{ssec:quality}
(216 items; $n{=}8$ per cell).
Mi = Mild, Mo = Moderate, Se = Severe.
Scores are on a 1 (low) to 5 (high) scale.
FA and Cr coincided on every fabrication/contradiction item in this
sample and are reported separately for completeness.}
\label{tab:llm_eval}
\centering
\footnotesize
\setlength{\tabcolsep}{2.5pt}
\renewcommand{\arraystretch}{1.02}
\begin{tabular}{cc*{9}{c}}
\toprule
\multirow{2}{*}{\textbf{Type}} &
\multirow{2}{*}{\textbf{Q}} &
\multicolumn{3}{c}{\textbf{EN}}
 &
\multicolumn{3}{c}{\textbf{KZ}} &
\multicolumn{3}{c}{\textbf{RU}} \\
\cmidrule(lr){3-5}
\cmidrule(lr){6-8}
\cmidrule(lr){9-11}
& &
\textbf{Mi} & \textbf{Mo} & \textbf{Se} &
\textbf{Mi} & \textbf{Mo} & \textbf{Se} &
\textbf{Mi} & \textbf{Mo} & \textbf{Se} \\
\midrule
\multirow{7}{*}{$\otimes$}
& AT & 2.88 & 4.25 & 5.00 & 4.00 & 4.25 & 4.88 & 3.88 & 4.50 & 5.00 \\
& Co & 1.75 & 2.00 & 1.50 & 2.12 & 2.25 & 1.88 & 1.88 & 2.50 & 1.25 \\
& Fl & 2.88 & 3.00 & 2.88 & 3.38 & 3.38 & 3.12 & 3.25 & 3.38 & 3.00 \\
& IC & 2.00 & 2.00 & 2.25 & 2.25 & 2.25 & 1.88 & 2.00 & 2.88 & 1.62 \\
& LC & 1.62 & 2.00 & 1.50 & 2.12 & 2.25 & 1.88 & 1.75 & 2.50 & 1.25 \\
& Re & 2.00 & 2.00 & 1.75 & 2.12 & 2.00 & 1.62 & 2.25 & 2.00 & 1.00 \\
& Sa & 3.88 & 4.25 & 5.00 & 4.00 & 4.25 & 4.88 & 4.00 & 4.50 & 5.00 \\
\midrule
\multirow{7}{*}{$\diamond$}
& AT & 4.00 & 4.12 & 5.00 & 2.62 & 4.12 & 5.00 & 4.25 & 4.25 & 5.00 \\
& Co & 4.00 & 3.75 & 2.38 & 4.25 & 3.88 & 3.00 & 4.38 & 4.00 & 3.38 \\
& Cr & 2.00 & 2.00 & 1.00 & 1.62 & 1.88 & 1.00 & 2.00 & 2.00 & 1.00 \\
& FA & 2.00 & 2.00 & 1.00 & 1.62 & 1.88 & 1.00 & 2.00 & 2.00 & 1.00 \\
& Fl & 4.00 & 3.88 & 2.88 & 4.25 & 4.00 & 3.75 & 4.38 & 4.00 & 4.00 \\
& Re & 4.00 & 3.38 & 2.25 & 2.12 & 3.50 & 2.62 & 4.25 & 3.25 & 2.62 \\
& Sa & 4.00 & 4.00 & 5.00 & 2.38 & 4.00 & 5.00 & 4.00 & 4.00 & 5.00 \\
\midrule
\multirow{7}{*}{$\neq$}
& AT & 4.25 & 4.88 & 5.00 & 4.25 & 4.88 & 5.00 & 4.75 & 5.00 & 5.00 \\
& Co & 3.38 & 2.50 & 2.12 & 3.88 & 3.88 & 3.25 & 3.38 & 3.12 & 3.38 \\
& Cr & 1.50 & 1.12 & 1.00 & 1.75 & 1.12 & 1.00 & 1.25 & 1.00 & 1.00 \\
& FA & 1.50 & 1.12 & 1.00 & 1.75 & 1.12 & 1.00 & 1.25 & 1.00 & 1.00 \\
& Fl & 3.62 & 3.12 & 2.62 & 4.00 & 4.00 & 3.75 & 3.75 & 3.75 & 4.00 \\
& Re & 3.50 & 2.62 & 2.12 & 3.75 & 3.25 & 2.38 & 3.38 & 2.75 & 2.88 \\
& Sa & 4.38 & 4.88 & 5.00 & 4.25 & 4.88 & 5.00 & 4.75 & 5.00 & 5.00 \\
\bottomrule
\end{tabular}
\vspace{0.4em}
\\[-0.2em]
{\footnotesize
\textbf{Legend:}
$\otimes$ = Context Inconsistency,
$\diamond$ = Fabrication,
$\neq$ = Contradiction.
AT = Adequacy of Type, Fl = Fluency, IC = Internal Consistency,
LC = Logical Coherence, Re = Relevance, Co = Coherence,
Cr = Credibility, FA = Factual Accuracy, Sa = Satisfaction.}
\end{table}

\section{\rev{Real-World Split Details}}
\label{app:realworld}

\rev{The 290 fake items were downloaded from the public factcheck.kz
archive in October 2024 (225 natively Russian, 65 natively Kazakh, after
removal of 15 exact duplicates from the initial Russian collection).
Each item was machine-translated into the other language with Google
Translate; a fluent bilingual author manually reviewed all 290
translations and judged all of them adequate, with no post-editing
required. Truthful negatives (290 per language) are original
human-written articles of the synthetic subset, selected via greedy
TF-IDF cosine matching to the fakes; they are excluded from the
train/dev splits of all models evaluated on the real-world data
(Appendix~\ref{app:training}). The faithful-rewrite condition
of Table~\ref{tab:confound} rewrites the same 290 Russian truthful
negatives with Gemini~2.0 Flash-Lite under a hallucination-free instruction
(preserve all facts, names, numbers, dates, and claims; introduce no
unsupported information; keep approximately the same length).}

\end{document}